\documentclass{article}

\usepackage{microtype}
\usepackage{graphicx}
\usepackage{subcaption}
\usepackage{booktabs} 

\usepackage{hyperref}

\usepackage[preprint]{icml2026}

\usepackage{amsmath}
\usepackage{amssymb}
\usepackage{mathtools}
\usepackage{amsthm}

\usepackage[capitalize,noabbrev]{cleveref}

\theoremstyle{plain}
\newtheorem{theorem}{Theorem}[section]

\theoremstyle{definition}

\newtheorem{assumption}[theorem]{Assumption}
\theoremstyle{remark}
\newtheorem{remark}[theorem]{Remark}

\usepackage[textsize=tiny]{todonotes}

\usepackage{caption}
\usepackage{amsmath, amssymb, amsfonts}

\usepackage[ruled,vlined,algo2e]{algorithm2e}

\usepackage{multicol}
\usepackage{multirow}

\usepackage{wrapfig}

\usepackage[utf8]{inputenc}
\usepackage[table]{xcolor} 
\usepackage{pifont} 
\usepackage[most]{tcolorbox}

\usepackage{pifont}  

\usepackage{enumitem}

\usepackage{mdframed}

\newmdenv[
  leftline=true,
  rightline=false,
  topline=false,
  bottomline=false,
  linecolor=gray,
  linewidth=2pt,
  backgroundcolor=gray!8,
  innerleftmargin=8pt,
  innerrightmargin=8pt,
  innertopmargin=6pt,
  innerbottommargin=6pt
]{insightbox}

\definecolor{checkgreen}{RGB}{76,175,80}   
\definecolor{crossorange}{RGB}{245,124,0}  
\newcommand{\cmark}{\textcolor{checkgreen}{\ding{51}}}   
\newcommand{\xmark}{\textcolor{crossorange}{\ding{55}}} 

\definecolor{headergray}{RGB}{230,230,240}
\definecolor{rowgray}{gray}{0.9}
\definecolor{rowgray1}{gray}{0.85}
\definecolor{rowgray2}{gray}{0.8}

\icmltitlerunning{Semi-supervised Meta Additive Model}

\begin{document}

\twocolumn[
\icmltitle{S$^2$MAM: Semi-supervised Meta Additive Model \\
for Robust Estimation and Variable Selection}


\icmlsetsymbol{equal}{*}

\begin{icmlauthorlist}
\icmlauthor{Xuelin Zhang}{a}
\icmlauthor{Hong Chen$^*$}{a}
\icmlauthor{Yingjie Wang}{b}
\icmlauthor{Tieliang Gong}{c}
\icmlauthor{Bin Gu}{d}
\end{icmlauthorlist}

\icmlaffiliation{a}{Huazhong Agricultural University}
\icmlaffiliation{b}{China University of Petroleum (East China)}
\icmlaffiliation{c}{Xi'an Jiaotong University}
\icmlaffiliation{d}{Jilin University}

\icmlcorrespondingauthor{Xuelin Zhang}{xlinml@163.com}
\icmlcorrespondingauthor{Hong Chen}{chenh@mail.hzau.edu.cn}

\icmlkeywords{Machine Learning, ICML}

\vskip 0.3in
]



\printAffiliationsAndNotice{}  

\begin{abstract}

Semi-supervised learning with manifold regularization is a classical framework for jointly learning from both labeled and unlabeled data, where the key requirement is that the support of the unknown marginal distribution has the geometric structure of a Riemannian manifold. Typically, the Laplace-Beltrami operator-based manifold regularization can be approximated empirically by the Laplacian regularization associated with the entire training data and its corresponding graph Laplacian matrix. However, the graph Laplacian matrix depends heavily on the prespecified similarity metric and may lead to inappropriate penalties when dealing with redundant or noisy input variables. To address the above issues, this paper proposes a new \textit{Semi-Supervised Meta Additive Model} (S$^2$MAM) based on a bilevel optimization scheme that automatically identifies informative variables, updates the similarity matrix, and simultaneously achieves interpretable predictions. Theoretical guarantees are provided for S$^2$MAM, including the computing convergence and the statistical generalization bound. Experimental assessments across 4 synthetic and 12 real-world datasets, with varying levels and categories of corruption, validate the robustness and interpretability of the proposed approach. 

\end{abstract}

\section{Introduction} \label{section1}

Manifold regularization provides an elegant and practical framework for developing semi-supervised learning (SSL) models by leveraging large amounts of unlabeled data alongside limited labeled data \citep{belkin2004semi,belkin2005manifold,yao2025manifold}.
The key assumption of manifold regularization is that the support of the intrinsic marginal distribution has the geometric structure of a Riemannian manifold \citep{johnson2007effectiveness,johnson2008graph}). Usually, the Laplace-Beltrami operator-based manifold regularization can be approximated empirically by the Laplacian regularization associated with the whole training data and the
corresponding similarity (adjacent) matrix \citep{belkin2004semi,roweis2000nonlinear}, where the similarity matrix is constructed 
by the principles of Gaussian fields and harmonic functions \citep{zhu2003semi} or the local and global consistency \citep{zhou2003learning}. Typical manifold regularization schemes include Laplacian regularized least squares (LapRLS) and Laplacian regularized support vector machine (LapSVM) \citep{belkin2006manifold}. Moreover, 
\citet{nie2010flexible} considered a flexible manifold embedding for semi-supervised dimension reduction, and \citet{qiu2018accelerating} further developed an accelerated version (called fast flexible manifold embedding, f-FME)  by reconstructing a smaller adjacency matrix with low-rank and sparse constraints. 

\begin{figure*}[!ht]
\centering
\includegraphics[width=0.8\textwidth]{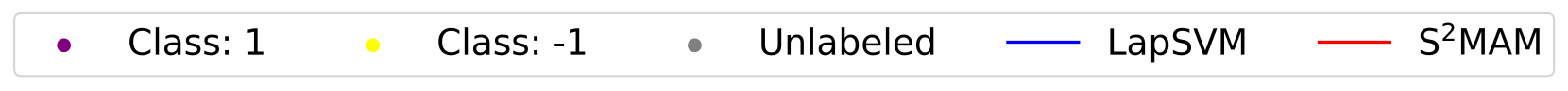}
\setcounter{subfigure}{0}
\begin{subfigure}[b]{0.24\textwidth}
\centering
\includegraphics[width=\textwidth]{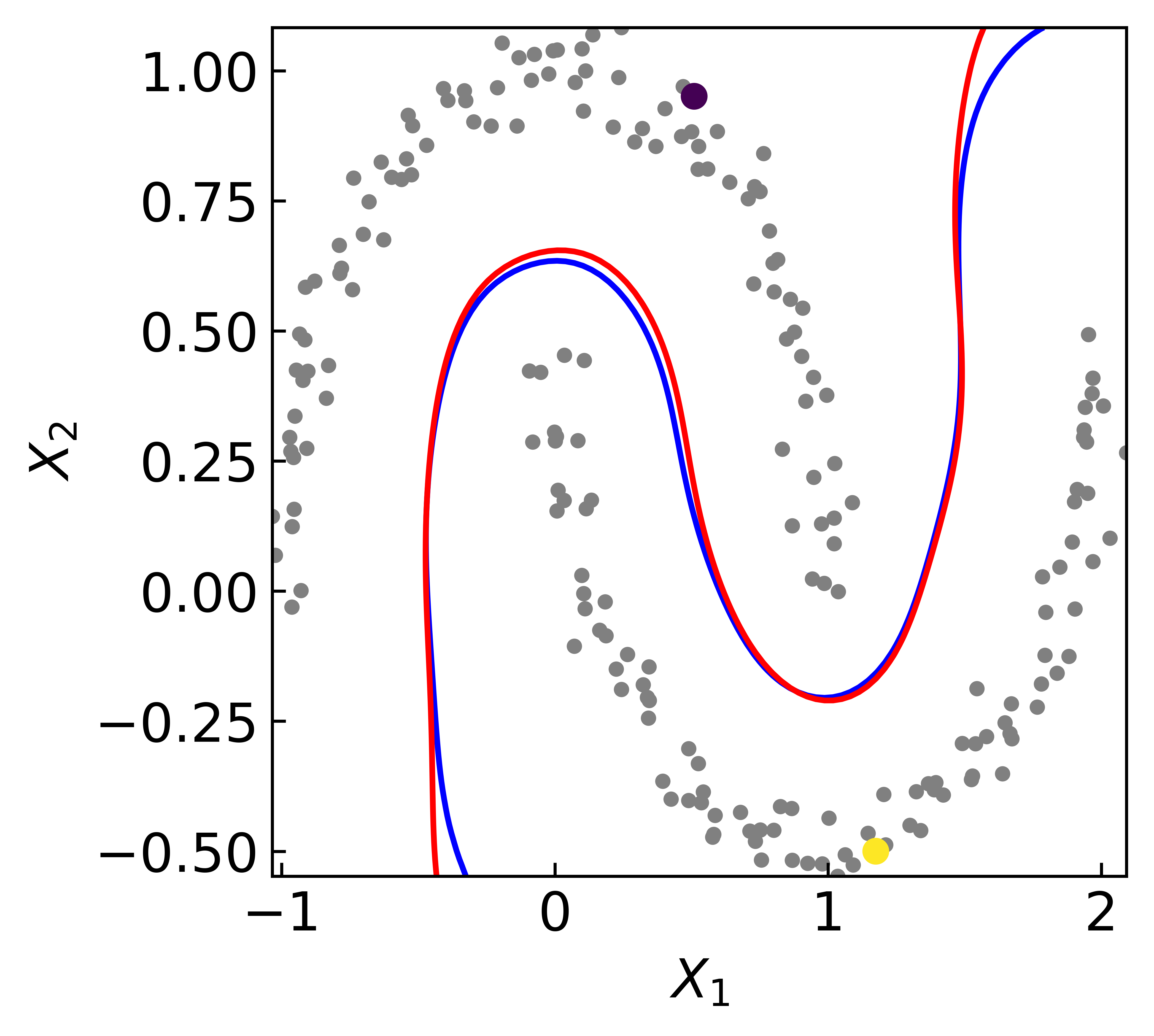}
\caption{Training on clean data}
\end{subfigure}
\begin{subfigure}[b]{0.24\textwidth}
\centering
\includegraphics[width=\textwidth]{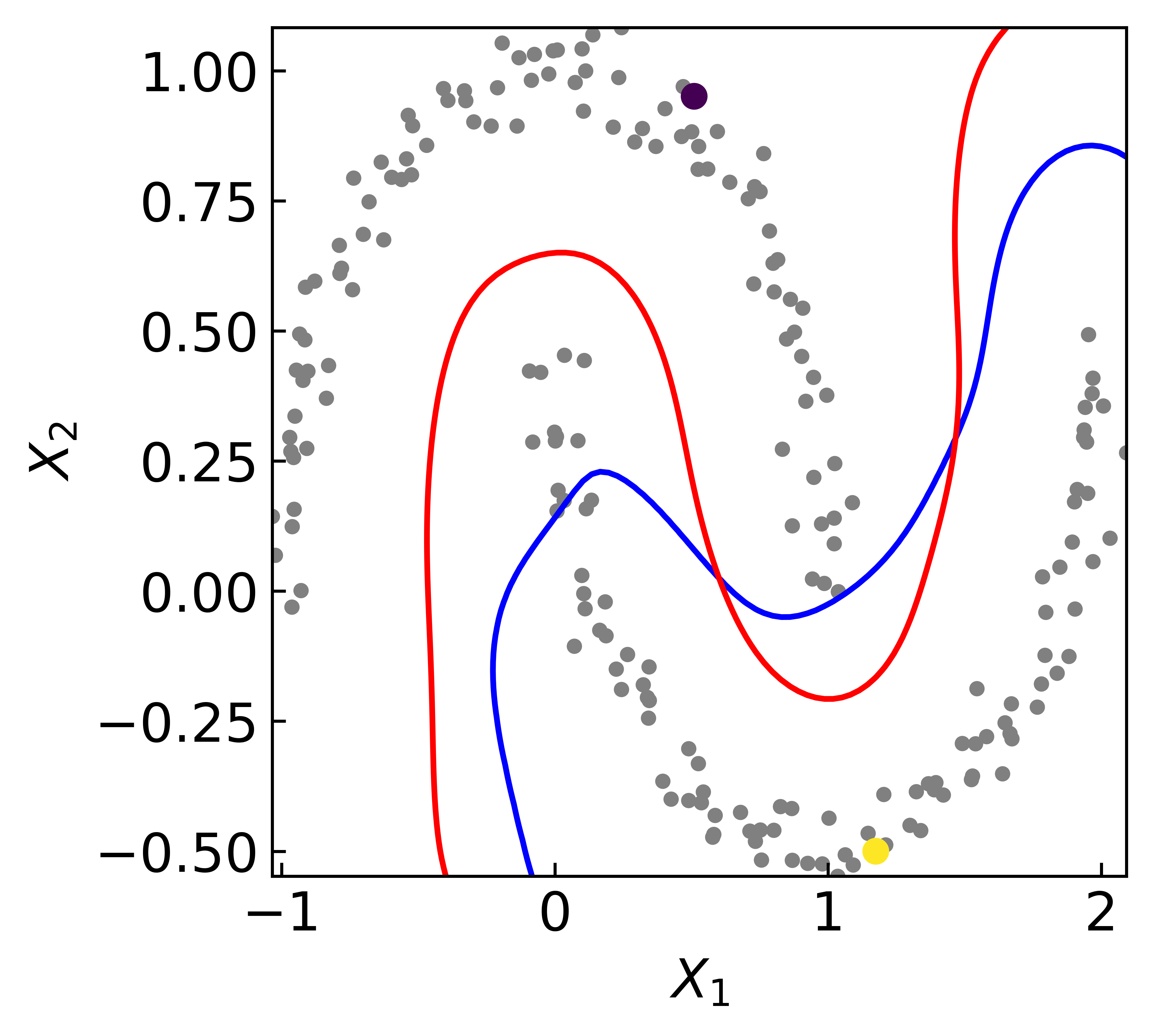}
\caption{LapSVM on noisy data}
\end{subfigure}
\begin{subfigure}[b]{0.24\textwidth}
\centering
\includegraphics[width=\textwidth]{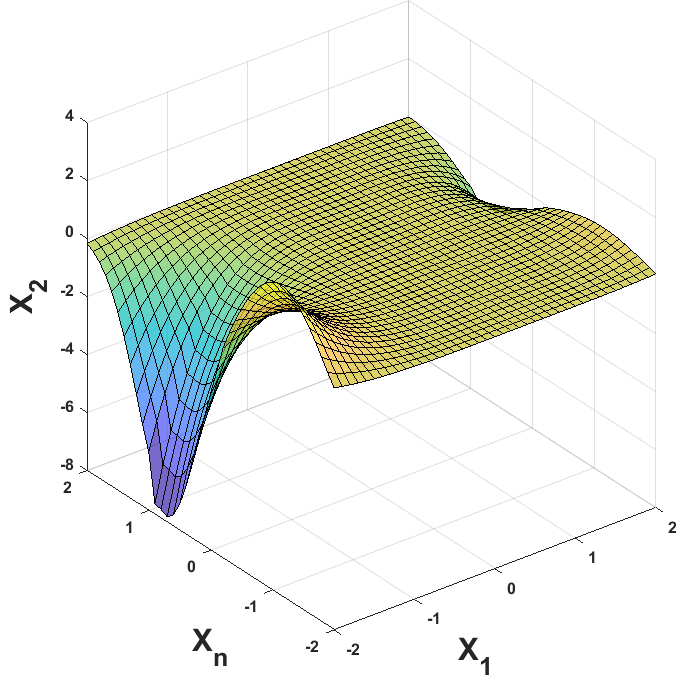}
\caption{LapSVM on noisy data}
\end{subfigure}
\begin{subfigure}[b]{0.24\textwidth}
\centering
\includegraphics[width=\textwidth]{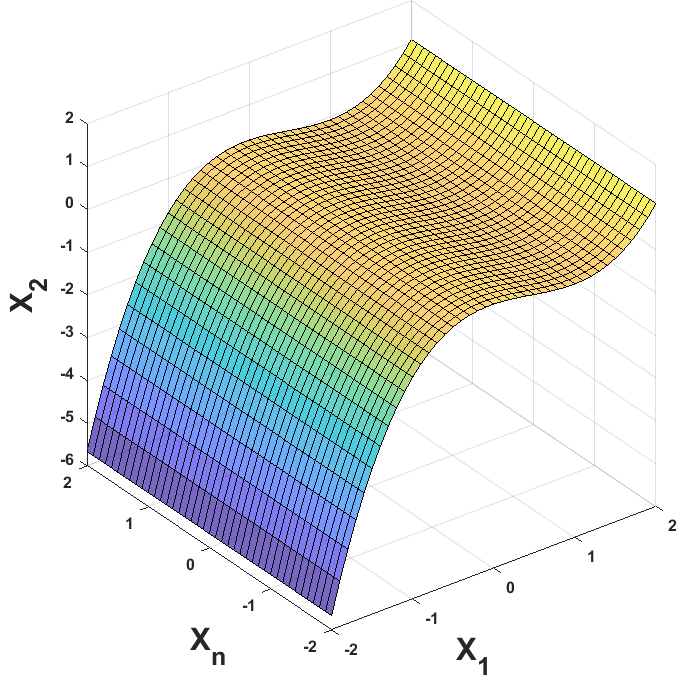}
\caption{S$^2$MAM on noisy data}
\end{subfigure}
\caption{Toy examples on the impact of noisy variables in the moon dataset for LapSVM and our S$^2$MAM.
(a) and (b) show the 2D prediction curves w.r.t the original input $X_1$ and $X_2$, where LapSVM is sensitive to feature corruptions $X_n$. (c) and (d) present the 3D decision surfaces on corrupted data, where S$^2$MAM is robust against the varying noisy variable $X_n$. 
The clean moon dataset contains inputs, $X_1$ and $X_2$. The corrupted data involves another noisy variable $X_n \in \mathcal{N}(100,100)$.
The used moon dataset contains 99 unlabeled points and only 1 labeled point per class.  
Please refer to \textit{Appendix} for detailed descriptions. 
}
\label{fig:toy_example}
\end{figure*}

Despite rapid progress, it remains difficult to validate the intrinsic manifold assumption \citep{belkin2004semi,johnson2008graph} across different types of data, e.g., data with redundant or even noisy variables.
Moreover, the investigation into the robustness and interpretability of manifold regularization is far behind its empirical applications, focusing only on prediction accuracy.
Existing manifold regularization models require prespecifying similarity matrices before semi-supervised training, leaving the adaptivity and robustness of manifold learning largely unexplored. For real applications, they unavoidably involve some abundant irrelevant and even noisy variables, and the pre-specified similarity metric associated with all the variables can not reflect the true adjacent relations properly. Uninformative and noisy variables often lead to significant deviations in the estimation of the manifold structure, seriously degrading the predictive capability of manifold regularization methods. 
As illustrated in Figure \ref{fig:toy_example}, the clean unlabeled data are beneficial to better fit the decision curve, while the randomly added noisy variables obviously hurt the performance of LapSVM (See \textit{Appendix} for detailed illustrations). The primary reason for the degraded performance is the computational bias in the similarity matrix, which directly affects the entire set of input variables \citep{nie2019semi,nie2021adaptive}.
This motivates the following open questions: 

\begin{tcolorbox}[colback=black!5!white, colframe=black!100!black, boxrule=0.5mm, left=0.1mm, right=0.1mm, top=0.1mm, bottom=0.1mm]
How to alleviate the impact of redundant and even noisy variables on SSL models with manifold regularization? 
How to design a new manifold regularization scheme that achieves robustness, interpretability, and prediction effectiveness simultaneously?
\end{tcolorbox}

Intuitively, we can address the above questions using a two-stage framework: first selecting informative variables (e.g., via Lasso \citep{Tibshirani-JRSS-1994}, SpAM \citep{Ravikumar-ICNIPS-2007}) and then applying manifold regularization with the refined input variables. However, this variable selection strategy is independent of the intrinsic manifold structure, and its accuracy cannot be guaranteed due to the scarcity of labeled data. Inspired by meta-learning for coreset selection \citep{borsos2020coresets, zhangtong2022coreset}, this paper considers assigning masks to all input variables, both labeled and unlabeled, retaining only the truly informative variables for modeling and constructing the similarity matrix. 

\begin{table*}[!t]
\centering
\caption{Properties of our S$^2$MAM and related models where ``SSL'' stands for semi-supervised learning. (D.D. and D.I. represent data-dependent and data-independent hypotheses. \cmark = satisfying the given information, and \xmark = not available for the information)}
\label{table1_highlight}
\resizebox{\textwidth}{!}{
\begin{tabular}{l|cccccccc}
\toprule
\rowcolor{headergray}
\textbf{Property} & \textbf{SpAM} & \textbf{LapRLS} & \textbf{f-FME} & \textbf{AWSSL} & \textbf{RER} & \textbf{SemiReward} & \textbf{PBCS} & \textbf{S$^2$MAM (ours)} \\
\midrule
\rowcolor{rowgray}
Learning Task & Supervised & SSL & SSL & SSL & SSL & Supervised & Supervised & \textbf{SSL} \\
Hypothesis Space &D.D. &D.I. &D.I. &D.I. &D.I. &D.I. &D.I. &D.D. \tabularnewline
\rowcolor{rowgray}
Optimization Framework & 1-level & 1-level & 1-level & 1-level & 1-level & 1-level & Bilevel & \textbf{Bilevel} \\
Interpretability & \cmark & \xmark & \xmark & \xmark & \xmark & \xmark & \xmark & \textbf{\cmark} \\
\rowcolor{rowgray}
Variable Selection & \cmark & \xmark & \cmark & \cmark & \cmark & \xmark & \xmark & \textbf{\cmark} \\
Noisy Variable Robustness & \xmark & \xmark & \xmark & \cmark & \cmark & \cmark & \xmark & \textbf{\cmark} \\
\rowcolor{rowgray}
Convergence Analysis & \xmark & \xmark & \xmark & \xmark & \xmark & \xmark & \cmark & \textbf{\cmark} \\
Generalization Analysis & \cmark & \xmark & \xmark & \xmark & \xmark & \xmark & \xmark & \textbf{\cmark} \\
\rowcolor{rowgray}
Comp. Complexity & \xmark & \xmark & \xmark & \xmark & \xmark & \xmark & \xmark & \textbf{\cmark} \\
\bottomrule
\end{tabular}}
\end{table*}

Nevertheless, there are several challenges along this way:
1) It is NP-hard to learn the discrete mask variables taking values in $\{0,1\}$ directly.
2) The bilevel optimization usually needs the computation on Hessian and Jacobian matrices, which leads to a heavy computation burden. 
3) Most kernel-based manifold regularization models construct the Gram matrix based on sample distance, which lacks the result's interpretability, e.g., screening the key variables associated with the response. 

\subsection{Contribution}
To address the aforementioned challenges, we incorporate meta-learning and sparse additive models into a manifold-regularized SSL framework and formulate a new \textit{Semi-Supervised Meta Additive Model} (S$^2$MAM) to enable automatic variable masking and sparse approximation for high-dimensional inputs, even in the presence of noisy variables. 

The core technique involves updating the decision function and the similarity matrix simultaneously, using appropriate masks on the input variables. The masks of S$^2$MAM are learned through a probabilistic meta-strategy.
Moreover, an efficient implementation is employed here to solve the bilevel optimization problem, which avoids the heavy computing burden on the implicit hypergradient calculation \citep{pedregosa2016hyperparameter}, Neumann series, and some variants with Hessian-vector or Jacobian-vector products \citep{ji2021bilevel,convergence_NIPS22}. 

The main contributions of this paper are summarized below:
\begin{itemize}
\item \textit{New statistical modeling}.
To the best of our knowledge, our S$^2$MAM is the first meta-learning method for manifold-regularized additive models, where a novel bilevel optimization scheme is formulated to achieve robust estimation and data-driven automatic variable selection simultaneously. 
By assigning flexible masks to individual variables, the proposed S$^2$MAM can reduce the impact of noisy variables on SSL tasks.

\item \textit{Computing and Theoretical Supports}.
An efficient probabilistic bilevel optimization is developed to additionally learn the discrete masks, utilizing both policy gradient estimation and the projection operation. This computational algorithm alleviates the computational burden of the discrete bilevel optimization framework and provides theoretical convergence guarantees.  Additionally, we establish upper bounds on the excess risk of the baseline S$^2$MAM model, implying that the proposed approach can achieve polynomial decay in generalization error. 

\item \textit{Empirical competitiveness}. 
Empirical results on several synthetic and real-world benchmarks demonstrate that the proposed S$^2$MAM can identify truly informative variables and achieve robust prediction even in the presence of redundant and noisy input variables.
	
\end{itemize}

\subsection{Comparisons with the Related Works}

\textit{Semi-supervised dimensionality reduction}. Recently, some efforts have been made to construct a flexible similarity matrix robust to feature corruptions for SSL with manifold regularization \citep{chen2018semi,nie2019semi,bao2024robust}. By rescaling the regression coefficients as variable weights, \citet{chen2018semi} developed an efficient SSL method to identify important variables, known as rescaled linear square regression. Another weighting approach in \citep{nie2019semi} is called auto-weighting semi-supervised learning (AWSSL), which adaptively assigns continuous weights on variables to update the similarity matrix. After dimension reduction, a specific classifier is employed for downstream tasks. 
A robust graph learning (RGL) method \citep{kang2019robust} combined label ranking regression and label propagation into a unified framework for weight graph construction and semi-supervised learning. Semi-supervised adaptive local embedding learning (SALE) \citep{nie2021adaptive} adaptively constructs two affinity graphs (based on labeled data and all embedding samples) separately to explore the local and global structures. \citet{bao2024robust} proposes an efficient model, robust embedding regression (RER), integrating a low-rank representation and Laplacian regularization.
Unlike these works, this paper considers the automatic assignment of discrete masks (0/1) to input features (variables) to screen for the truly active variables.

\textit{Sparse additive models}. Additive models \citep{Stone-A.S-1985,hastie}, as natural nonparametric extensions of linear models, have been burgeoning in high-dimensional data analysis due to their attractive properties, i.e., overcoming the curse of dimensionality, the flexibility of function approximation, and the ability of variable selection \citep{Meier-A.S-2009, Christmann-Computstat.dataan-2012-additive, Yuan-Ann.Stat-2016-additive,chen-2020-tnnls-spmam}.
In recent years, many sparse additive models have been proposed for various theoretical or empirical motivations, see e.g., \citep{lV-Ann.Stat-2017-additive,haris2022generalized,bouchiat2023improving,duong2024cat}.
Naturally, the paradigm of additive models can be applied to semi-supervised learning settings. As far as we know, there are only three papers that touched on this topic \citep{culp2008iterative,culp2009sslam,culp2011propagated}. However, not all of them consider the robustness of manifold learning to noisy variables and ignore data-driven discovery of variable structure. 
These stringent restrictions on the predefined similarity matrix and variable structure may result in a severe degradation of existing models under complex noise conditions.

\textit{Meta learning for sample/variable selection.} The meta-based masking policy was developed in \citep{borsos2020coresets}, where a bilevel neural network is designed for automatic supervised coreset selection. Furthermore, its improved version with probabilistic bilevel optimization is proposed for supervised classification \citep{zhangtong2022coreset}, especially for corrupted and imbalanced data. Indeed, \citet{zhangtong2022coreset} also provides an example of variable selection, while it is limited to the supervised learning case and doesn't concern the impact of noisy variables. To the best of our knowledge, no prior effort has explored meta-based masking policies for semi-supervised additive models.

Table \ref{table1_highlight} provides a comprehensive comparison on the properties of S$^2$MAM with recent state-of-the-art models, including SpAM \citep{Ravikumar-ICNIPS-2007}), LapRLS \citep{belkin2006manifold}), f-FME \citep{qiu2018accelerating}, AWSSL \citep{nie2019semi}, RER \citep{bao2024robust}, SemiReward \citep{li2024semireward}, and the probabilistic bilevel coreset selection (PBCS) \citep{zhangtong2022coreset}. 
S$^2$MAM offers a distinct advantage in maintaining robustness against complex noise. The inherent data-dependency hypothesis poses substantial challenges for theoretical analysis, particularly for deriving generalization bounds relevant to the underlying data distribution.

\section{Semi-supervised Additive Models} \label{section2}

This section first introduces a manifold-regularized semi-supervised additive model \citep{culp2011propagated} as the basic model, and then formulates the S$^2$MAM within a discrete bilevel optimization framework. 
Furthermore, a probabilistic bilevel scheme solves the NP-hard discrete optimization problem. 

\subsection{Revisiting Manifold Regularized Sparse Additive Model} \label{sec:additive}

Let $\mathcal{X} = \{\mathcal{X}^{(1)}, \cdots, \mathcal{X}^{(p)}\} \in \mathbb{R}^p$ be a compact input space and the output space $\mathcal{Y}\in\mathbb{R}$.
Denote $\rho$ as the joint distribution on $\mathcal{X} \times \mathcal{Y}$, and $\rho_\mathcal{X}$ as the marginal distribution with respect to $\mathcal{X}$ induced by $\rho$. 
The training set $\mathbf{z} = \{\mathbf{z}_l, \mathbf{z}_u\}$ involves the labeled set $\mathbf{z}_l= \{(x_i,y_i)\}_{i=1}^l$ and the unlabeled set $\mathbf{z}_u=\{x_i\}_{i=l+1}^{l+u}$, where each input $x_i=(x_i^{(1)}, \cdots, x_i^{(p)})^T\in\mathbb{R}^p$ with $x_i^{(j)}\in \mathcal{X}^{(j)}$ and output $y_i\in\mathbb{R}$. The hypothesis space of additive models can be formulated as $\mathcal{F}=\{f: f(x)=\sum_{j=1}^p f^{(j)}(x^{(j)}), f^{(j)} \in \mathcal{F}^{(j)}\},$
where $x^{(j)} \in \mathcal{X}^{(j)}$ and $\mathcal{F}^{(j)}$ is the component function space on $\mathcal{X}^{(j)}$ \citep{Ravikumar-ICNIPS-2007}. 
Typical candidates of additive hypothesis space include the basis expansion space \citep{Meier-A.S-2009}, the reproducing kernel Hilbert space (RKHS) \citep{Raskutti-JMLR-2012-additive,wang2023tspam}, and the network-based space \citep{Agarwal-2020, duong2024cat}.

This paper chooses $\mathcal{H}_{K^{(j)}}$ to form the additive hypothesis space, where $\mathcal{H}_{K^{(j)}}$ is the RKHS associated with Mercer kernel $K^{(j)}$ defined on $\mathcal{X}^{(j)} \times \mathcal{X}^{(j)}, j \in\{1, \ldots, p\}$. Equipped by component function $f^{(j)}: \mathcal{X}^{(j)}\rightarrow\mathbb{R}, j \in\{1, \ldots, p\}$, the additive hypothesis space can be further defined as
\begin{equation}
\mathcal{H}=\Big\{f=\sum_{j=1}^p f^{(j)}: f^{(j)} \in \mathcal{H}_{K^{(j)}}, 1 \leq j \leq p\Big\}
\end{equation}
with $\|f\|_K^2= \inf \left\{\sum_{j=1}^p\|f^{(j)}\|_{K^{(j)}}^2: f=\sum_{j=1}^p f^{(j)}\right\}$.
Indeed, $\mathcal{H}$ is an RKHS associated with kernel $K=\sum_{j=1}^p K^{(j)}$.

Due to the Representer Theorem of RKHS \citep{smola1998learning}, the prediction function of supervised additive models in RKHS often provides a parameterized representation \citep{Yuan-Ann.Stat-2016-additive} as
\begin{equation} \label{additive_prediction} 
f(\cdot)=\sum_{j=1}^p \sum_{i=1}^l \alpha_i^{(j)}{K}_i^{(j)}(x_i^{(j)},\cdot).
\end{equation}

Given a predictor $f:\mathcal{X}\rightarrow\mathbb{R}$, denote 
$\mathbf{f}=(f(x_{1}), \ldots, f(x_{l+u}))^T$
as the prediction vector associated with the labeled data $\mathbf{z}_l$ and the unlabeled data $\mathbf{z}_u$. 
Let $\lambda_1,\lambda_2>0$ be the regularization coefficients and let $\tau_j$ be the positive weight to different input variables for $j=1,\cdots,p$.
Then, the additive model for regularized Laplacian regression can be formulated as 
\begin{equation}\label{f_z}
	f_{\mathbf{z}}=\underset{f \in \mathcal{H}}{\arg \min }\left\{\mathcal{E}_{\mathbf{z}}(f)+\lambda_{1} \Omega_{\mathbf{z}}(f)+\frac{\lambda_{2}}{(l+u)^{2}} \mathbf{f}^{T} \boldsymbol{L} \mathbf{f}\right\},
\end{equation}
where empirical risk $\mathcal{E}_{\mathbf{z}}(f)=\frac{1}{l} \sum_{i=1}^{l}(y_{i} - f(x_i))^{2},$
the sparse regularization $\Omega_{\mathbf{z}}(f)$ is formulated by the $\ell_{2,1}$ scheme of $\inf_{{\alpha}^{(j)}} \Big\{\sum_{j=1}^p \tau_{j}\|\boldsymbol{\alpha}^{(j)}\|_2: f=\sum_{j=1}^p \sum_{i=1}^l\alpha_i^{(j)}{K}_i^{(j)}(x_i^{(j)},\cdot)\Big\},$
and the term $\mathbf{f}^{T} \boldsymbol{L} \mathbf{f}$ is the manifold regularization \citep{culp2011propagated}. Here, 
$\boldsymbol{L}=\boldsymbol{D}-\boldsymbol{W}$ is the graph Laplacian, and
diagonal matrix $\boldsymbol{D}$ satisfies
$D_{ii} = \sum_{j=1}^{l+u} W_{ij}$ and $W_{ij}$ is the adjacent weight for inputs $x_i$ and $x_j$, e.g., $W_{ij}= \exp\{-\|x_i-x_j\|_2^2 / \mu^2\}$ with bandwidth $\mu$.

\begin{remark}
If the $j$-th variable is not truly informative, $\boldsymbol{\alpha}_{\mathbf{z}}^{(j)}=(\alpha_{\mathbf{z}, 1}^{(j)}, \ldots,\alpha_{\mathbf{z},l+u}^{(j)})^T \in \mathbb{R}^{l+u}$ is expected to satisfy $\|\boldsymbol{\alpha}_{\mathbf{z}}^{(j)}\|_2=\sqrt{\sum\limits_{i=1}^{l+u}\left|\alpha_{\mathbf{z}, i}^{(j)}\right|^2}=0$. Thus, $\ell_{2, 1}$-regularizer is employed as the penalty. Obviously, noisy input variables may bring an inappropriate similarity matrix $\boldsymbol{W}$. Naturally, it is necessary to improve the robustness of (\ref{f_z}) against noisy variables by replacing the pre-specified similarity measure (i.e., $\boldsymbol{W},\boldsymbol{L}$) in manifold regularization with an adaptive masking strategy. 
\end{remark}

\subsection{Discrete Bilevel Framework for S$^2$MAM} \label{sec_2.2}

To mitigate the negative impact of noisy variables on Laplacian regularization in (\ref{f_z}), we introduce a bilevel optimization framework for automatically learning variable masks. In particular, both the decision function $f$ and Laplacian matrix $\boldsymbol{L}$ are updated by the learned masks.

Denote $\ell(\cdot)$ as the loss function, $f(x;\boldsymbol{\alpha})$ as a decision function in RKHS $\mathcal{H}$ with spanning parameter $\boldsymbol{\alpha}$ and the mask $\boldsymbol{m} \in\{0,1\}^{p}$ as a binary vector, where $m_i=1$ implies $i$-th variable is selected as the informative one and otherwise ignored. {$\alpha$ denotes the coefficient parameter of the additive model. The bilevel framework for directly learning the discrete masks is formulated as follows.

\textbf{Upper Level:}
Given the meta dataset $D_{meta}=\{(x_i, y_i)\}_{i=1}^l$, we formulate the discrete optimization
\begin{equation}
\min _{\boldsymbol{m} \in \tilde{\mathcal{C}}} \mathcal{L}\left(\hat{\boldsymbol{\alpha}}(\boldsymbol{m})\right) = \frac{1}{l} \sum_{i=1}^l \ell\left(f(x_i ; \hat{\boldsymbol{\alpha}}(\boldsymbol{m})), y_i\right),
\end{equation}
where the mask $\boldsymbol{m}$ is the learnable parameter in the upper level, $\boldsymbol{\alpha}$ is the parameter of the decision function in the lower level depending on $\boldsymbol{m}$, and $\tilde{\mathcal{C}}=\left\{\boldsymbol{m}: m_i\in\{0, 1\},\|\boldsymbol{m}\|_0 \leq C, i=1,2,\cdots,p \right\}$ is the feasible region of $\boldsymbol{m}$ with the size of selected variables $C$. 

\textbf{Lower Level:}
Based on the whole training set $D_{total}$ involving $D_{meta}$ and unlabeled samples $\{x_i\}_{i=l+1}^{l+u}$, the predictor of lower level optimization problem is 
\begin{equation} \label{s2mam_prediction} 
\begin{aligned}
\hat{f}(x)=&\sum_{j=1}^p \hat{f}^{(j)} (m_j x^{(j)}) \\
= &\sum_{j=1}^p \sum_{i=1}^{l+u} \alpha_i^{(j)}{K}_i^{(j)}(m_j x_i^{(j)}, m_j x^{(j)}),
\end{aligned}
\end{equation}
where $\hat{\boldsymbol{\alpha}}=\underset{\boldsymbol{\alpha}\in \mathbb{R}^{(l+u)\times p}}{\arg \min } \mathcal{R}(\boldsymbol{\alpha};\boldsymbol{m};\boldsymbol{L})$,
with risk defined by $\mathcal{R}(\boldsymbol{\alpha};\boldsymbol{m};\boldsymbol{L}) = \frac{1}{l} \sum_{i=1}^l \ell(f(x_i\odot\boldsymbol{m}; \boldsymbol{\alpha}), y_i) +\lambda_1\sum_{j=1}^p \tau_j\|\boldsymbol{\alpha}^{(j)}\|_2 + \frac{\lambda_{2}}{(l+u)^{2}} \mathbf{f}^{T} \boldsymbol{L} \mathbf{f}.$

Different from (\ref{f_z}), the Laplacian matrix $\boldsymbol{L}$ is computed based on the masked similarity matrix $\boldsymbol{W}$ with measure function $\mathcal{W}(\cdot,\cdot)$ and element ${W}_{ij}=\mathcal{W}(x_i\odot \boldsymbol{m},x_j\odot \boldsymbol{m}), i,j\in\{1,2,\cdots,l+u\}.$

Usually, it is intractable to solve the above discrete bilevel problem directly. Fortunately, we can formulate its continuous probabilistic form using policy gradient estimation \citep{zhangtong2022coreset} and develop an efficient gradient-based optimization algorithm as follows.

\subsection{Probabilistic Bilevel Framework for S$^2$MAM} \label{sec_ps2mam}

It is popular to transform the discrete tuning parameter space into the continuous probability space for bilevel optimization \citep{zhaoTNN23pmwn}. For simplicity, $m_i$ can be considered as a Bernoulli random variable $m_i \sim \operatorname{Bern}\left(s_i\right)$, where $s_i \in [0,1]$ represents the probability of $m_i=1$. 
Denote the domain on probability variable $\boldsymbol{s}=(s_1,...,s_p)\in\mathbb{R}^p$ as 
\begin{equation}
\mathcal{C}=\left\{\boldsymbol{s}: 0 \preceq s_i \preceq 1, \|\boldsymbol{s}\|_1 \leq C,  i=1,2,\cdots, p\right\}.
\end{equation}
The discrete bilevel optimization in Section \ref{sec_2.2} can be relaxed into the following expected form
\begin{equation} \label{probabilistic_bilevel}
\begin{aligned} 
&\min_{\boldsymbol{s} \in \mathcal{C}} \text{ }\Phi(\boldsymbol{s})=\mathbb{E}_{p(\boldsymbol{m} \mid \boldsymbol{s})} \mathcal{L}\left(\boldsymbol{\alpha}^*(\boldsymbol{m})\right), \\
&\text{ s.t. }~\boldsymbol{\alpha}^*(\boldsymbol{m}) \in \underset{\boldsymbol{\alpha}\in \mathbb{R}^{(l+u)\times p}}{\arg \min } \mathcal{R}(\boldsymbol{\alpha};\boldsymbol{m};\boldsymbol{L}).
\end{aligned}
\end{equation}

\begin{remark}
Under the independent assumption on variable $m_i$, we can derive its continuous distribution $p(\boldsymbol{m} \mid \boldsymbol{s})=\Pi_{i=1}^p \left(s_i\right)^{m_i}\left(1-s_i\right)^{\left(1-m_i\right)}$. Since $\mathbb{E}_{\boldsymbol{m} \sim p(\boldsymbol{m}\mid\boldsymbol{s})}\|\boldsymbol{m}\|_0=\sum_{i=1}^p s_i$, the original domain $\tilde{\mathcal{C}}=\left\{\boldsymbol{m}: m_i\in\{0, 1\},\|\boldsymbol{m}\|_0 \leq C, i=1,2,\cdots,p \right\}$ is transformed into $\mathcal{C}$ on probability $\boldsymbol{s}$. 
While we assume independent Bernoulli variables for $m_i$ to facilitate efficient policy gradient estimation, S$^2$MAM implicitly handles feature dependencies through the shared manifold structure $L$ in the lower-level problem.
\end{remark}

\subsection{Computing Algorithm of S$^2$MAM}
Initialize the decision parameter $\boldsymbol{\alpha}^0 = \boldsymbol{0}$, mask  $\boldsymbol{m}^0 = \boldsymbol{1}$, probability  $\boldsymbol{s}^0 = \frac{C}{p} \cdot \boldsymbol{1}$ and select Laplacian matrix associated with original $(x_1,\cdots,x_{l+u})$ as $\boldsymbol{L}^0$. Before each iteration, a sample batch $\mathcal{B}$ is selected from the whole training set.
The computing steps of probabilistic S$^2$MAM are summarized in Algorithm 1. The procedures for solving (\ref{probabilistic_bilevel}) at the $t$-th iteration contain: 

\textbf{Step 1: Computing $\boldsymbol{\alpha}^{t}$ with $\boldsymbol{m}^{t-1}$ and $\boldsymbol{L}^{t-1}$}
\begin{equation}
\boldsymbol{\alpha}^{t}=\underset{\boldsymbol{\alpha}\in \mathbb{R}^{(l+u)\times p}}{\arg \min } \mathcal{R}(\boldsymbol{\alpha};\boldsymbol{m}^{t-1};\boldsymbol{L}^{t-1}),
\end{equation}
with risk defined by $\mathcal{R}(\boldsymbol{\alpha}^{t-1};\boldsymbol{m}^{t-1};\boldsymbol{L}^{t-1})$. The computing algorithm for Step 1, based on the alternating direction method of multipliers, is presented in \textit{Appendix}.

\textbf{Step 2: Computing $\boldsymbol{s}^{t}$ and $\boldsymbol{m}^{t}$ with $\boldsymbol{\alpha}^{t}$.}
From the probabilistic S$^2$MAM in (\ref{probabilistic_bilevel}), the learning target changes from the discrete masks $\boldsymbol{m}$ into the continuous probability $\boldsymbol{s}$, which is updated by the policy gradient estimator \citep{zhangtong2022coreset} as $\nabla_{\boldsymbol{s}} \Phi(\boldsymbol{s}) = \mathbb{E}_{p(\boldsymbol{m} \mid \boldsymbol{s})} \mathcal{L}\left(\boldsymbol{\alpha}^*(\boldsymbol{m})\right) \nabla_{\boldsymbol{s}} \ln p(\boldsymbol{m}\mid\boldsymbol{s}).$
This computational procedure provides unbiased gradient estimation without imposing a heavy computational burden on the inverse Hessian or on implicit differentiation. 	

Denote $\eta^t$ as the step size for updating the upper level parameter $\boldsymbol{s}$ at the $t$-th step. 
Given $\boldsymbol{\alpha}^{t}$, $\boldsymbol{s}$ can be updated by the projected stochastic gradient descent below 
\begin{equation} \label{step_2}
\boldsymbol{s}^{t} \leftarrow \mathcal{P}_{\mathcal{C}}\left(\boldsymbol{s}^{t-1}-\eta^t \mathcal{L}\left(\boldsymbol{\alpha}^{t}\right) \nabla_{\boldsymbol{s}} \ln p(\boldsymbol{m}^{t-1} \mid \boldsymbol{s}^{t-1})\right),
\end{equation}
where the projection $\mathcal{P}_{\mathcal{C}}(\boldsymbol{s})$ from $\boldsymbol{s}$ to the domain $\mathcal{C}$ is summarized in Algorithm 2 in Appendix. Then,  $\boldsymbol{m}^{t} = (m_1^t,\cdots,m_{p}^t) \in \mathbb{R}^p$ follows from Bernoulli distribution, where $m_i^t \sim \operatorname{Bern}\left(s_i^t \right)$. \textit{Appendix} proves the closed-form solution in the projection computation.

\textbf{Step 3: Updating Laplacian matrix $\boldsymbol{L}^{t}$ with $\boldsymbol{m}^{t}$}
\begin{equation} \label{step_3}
	\boldsymbol{L}^t = \boldsymbol{D}^t - \boldsymbol{W}^t, 
\end{equation}
where the diagonal matrix $\boldsymbol{D}^t\in \mathbb{R}^{(l+u)\times(l+u)}$ satisfies $D^t_{i i}=\sum_{j=1}^{l+u} W_{ij}$, and $W_{ij}= \exp\{-\|x_i\odot\boldsymbol{m}^{t}-x_j\odot\boldsymbol{m}^{t}\|_2^2 / \mu^2\}$ with the bandwidth parameter $\mu>0$. The metric $W_{ij}$ evaluates the similarity between samples $x_i$ and $x_j$ that share the same mask $\boldsymbol{m}^{t}$. 
Specifically, we employ the expected mask $\mathbb{E}[m]=s$ to ensure deterministic and consistent predictions on unseen data.
Finally, we obtain the decision function in (\ref{s2mam_prediction}) with coefficient $\boldsymbol{\alpha}$ and mask $\boldsymbol{m}$.

To mitigate the quadratic cost of the graph Laplacian and probabilistic mask optimization on high-dimensional or large-size datasets, we've adopted two efficient strategies for acceleration, including preprocessing high-dimensional inputs (e.g., large images) via a pretrained CNN to extract a low-dimensional embedding, replaced by Random Fourier Features (RFF), which reduces complexity from $\mathcal{O}((l+u)^2)$ to $\mathcal{O}((l+u)D)$ where $D\ll l+u$.

\section{Theoretical Assessments} \label{section3} 

For the proposed S$^2$MAM, this section presents its computational convergence and generalization analysis for the basic model (\ref{f_z}) in Section \ref{sec:additive}. 
Corresponding proofs with computing complexity analysis are left in \textit{Appendix}.

\subsection{Computing Convergence Analysis}

We now establish the theoretical guarantee of convergence to the optimal policy for policy gradient estimation in Step 2. The following assumption has been widely used to characterize the convergence behavior of projection-based algorithms \citep{pedregosa2016hyperparameter} and bilevel optimization with minibatches \citep{shu2019meta}.

\begin{assumption} \label{assumption_convergence}
Denote $\mathcal{L}_{\mathcal{B}}$ as the loss on selected batch $\mathcal{B}$. Assume that $\Phi(\boldsymbol{s})$ is $L$-smooth, constant $\sigma>0$, there hold 
$\mathbb{E}[\mathcal{L}_{\mathcal{B}}\left(\boldsymbol{\alpha}^*(\boldsymbol{m})\right) \nabla_{\boldsymbol{s}} \ln p(\boldsymbol{m} \mid \boldsymbol{s}^t)-\nabla_{\boldsymbol{s}} \Phi(\boldsymbol{s}^t)] =0$, and $\mathbb{E}\left\|\mathcal{L}_{\mathcal{B}}\left(\boldsymbol{\alpha}^*(\boldsymbol{m})\right) \nabla_{\boldsymbol{s}} \ln p(\boldsymbol{m} \mid \boldsymbol{s}^t)-\nabla_{\boldsymbol{s}} \Phi(\boldsymbol{s}^t)\right\|^2 \leq \sigma^2$.
\end{assumption}

\begin{theorem} \label{convergence_theorem}
At the $t$-th iteration, let the step size $\eta^t = \frac{c}{\sqrt{t}}\leq \frac{1}{L}$ for some constant $c>0$, and denote the gradient mapping $\mathcal{G}^t=\frac{1}{\eta^t}\left(s^t-\mathcal{P}_{\mathcal{C}}\left(s^t-\eta^t \nabla_s \Phi\left(s^t\right)\right)\right)$.
Under Assumption \ref{assumption_convergence}, there holds
\begin{equation*}
\min_{1\leq t\leq T}  \mathbb{E}\left\|{\mathcal{G}}^t\right\|^2  \lesssim \mathcal{O}\left(T^{-\frac12}\right).
\end{equation*}
\end{theorem}

\begin{remark}
Indeed, \citet{zhangtong2022coreset} demonstrates that the average gradient $\frac{1}{T} \sum_{t=1}^T \mathbb{E}\left\|\mathcal{G}^t\right\|^2$ of the policy gradient estimation converges to a small constant as $T \rightarrow \infty$. With the help of refined step size $\eta^t=\frac{c}{\sqrt{t}}$, our results in Theorem \ref{convergence_theorem} shows better convergence property w.r.t. $T$. The empirical and theoretical analysis of algorithmic computation complexity is left in \textit{Appendix}.
\end{remark}

\subsection{Generalization Error Analysis}
The expected risk of $f: \mathcal{X} \rightarrow \mathcal{Y}$, w.r.t.  $\mathcal{E}_{\mathbf{z}}(f)$ in (\ref{f_z}), is measured by $\mathcal{E}(f)=\int_{\mathcal{X} \times \mathcal{Y}} (f(x)-y)^2 d \rho(x,y).$
It is well known that $f_\rho=\int_{\mathcal{Y}} y d \rho(y|\cdot)$
is the minimizer of $\mathcal{E}(f)$ over all measurable functions, where $\rho(y|x)$ denotes the conditional distribution of $y$ for given $x$. 
This work describes how fast $f_{\mathbf{z}}$ defined in (\ref{f_z}) approximates $f_\rho$ as the sample size increases. To the best of our knowledge, this is the first theoretical endeavor to characterize the generalization behavior of semi-supervised additive models.

Before presenting our results, we recall the necessary assumptions and definitions used here, which have been widely employed to bound the excess risk in supervised paradigms \citep{christmann2016learning,chen-2020-tnnls-spmam} and SSL models \citep{liu2018generalization,chen2018semi}.

\begin{assumption} \label{bounded_M1} (\cite{christmann2016learning})
For any $x\in\mathcal{X}$, there exists some $M \geq 0$ such that $\rho(\cdot\mid x)$ is almost everywhere supported on $[-M, M]$. Assume $f_{\rho}=\sum_{j=1}^pf_{\rho}^{(j)}$ with $0<r \leq \frac{1}{2}$ and $f_{\rho}^{(j)}=L_{K^{(j)}}^r\left(g_j^*\right)$ with some $g_j^* \in L_2(\rho(\mathcal{X}^{(j)}))$ for any $j \in\{1, \ldots, p\}$, where $L_2(\rho(\mathcal{X}^{(j)}))$ is the square-integrable space on  $\mathcal{X}^{(j)}$ and $L_{K^{(j)}}^r$ is $r$-power of integral operator $L_{K^{(j)}}: L_2(\rho(\mathcal{X}^{(j)})) \rightarrow L_2(\rho(\mathcal{X}^{(j)}))$ associated with kernel $K^{(j)}$.
\end{assumption}

\begin{assumption} \label{assumption_bounded_w}
Each entry of the similarity matrix $\boldsymbol{W}$ satisfies $0\leq W_{ij} \leq w$ for a positive constant $w$.
\end{assumption}

\begin{assumption} \label{kernel}
Let $C^{{v}}$ be a $\nu$-times continuously differentiable function set. Assume that $K^{(j)} \in C^\nu\left(\mathcal{X}^{(j)} \times \mathcal{X}^{(j)}\right), j \in\{1, \ldots, p\}$. 
\end{assumption} 

Define $\pi(f)(x)=\max\{\min\{f(x),M\},-M\}, \forall f\in\mathcal{H}$, as truncated output under Assumption \ref{bounded_M1}. This truncated operator has been used extensively for error analysis of learning algorithms, see e.g., \citep{Shi-JMLR-2019-sparsekernel}. Since $\mathcal{E}(\pi(f)) \leq \mathcal{E}(f)$ for any $f\in\mathcal{H}$, here we state the upper bound of $\mathcal{E}\left(\pi\left(f_{\mathbf{z}}\right)\right)-\mathcal{E}\left(f_{\rho}\right)$ to get a tighter generalization characterization for the manifold regularized additive model in (\ref{f_z}). 

\begin{theorem} \label{generalization bound}
Let $\lambda_1=(l+u)^{-\Delta}$, $\lambda_2=\lambda_1^{1-r}$ for some $\Delta>0$ and $0<r\leq1/2$. Under Assumptions \ref{bounded_M1}-\ref{kernel}, for any $0<\delta<1/2$, there holds
\[
\resizebox{0.49\textwidth}{!}{$
\begin{aligned}
\mathcal{E}\left(\pi\left(f_{\mathbf{z}}\right)\right)-\mathcal{E}\left(f_{\rho}\right)
\lesssim \log(\frac{8}{\delta}) p \left(\mathcal{O}\left((l+u)^{-\Theta} \right)
+\mathcal{O}\left(l^{-1/2}\right)\right) ,
\end{aligned}
$}
\]
with confidence at least $1-2\delta$, 
where $\Theta = \min\{\Delta r, 1+\Delta(r-1),\Delta(5r/2-3/2)+1/2, 2/(2+\zeta), 3/2-\Delta r, 1/2\}$ with
$\zeta= \begin{cases}\frac{2}{1+2 v}, & v \in(0,1] \\ \frac{2}{1+v}, & v \in(1,3 / 2] \\ \frac{1}{v}, & v \in(3 / 2, \infty)\end{cases}$.
\end{theorem}

\begin{remark}
The term $(l+u)^{-\Theta}$ characterizes the benefit of manifold regularization using unlabeled data. As $u \to \infty$, this term decays faster than the supervised error $l^{-1/2}$, effectively reducing the generalization gap by leveraging the geometric structure of the entire data distribution.
Besides the additional advantage of the interpretability of input variables, the basic model (\ref{f_z}) of S$^2$MAM also achieves the polynomial decay rate of excess risk, which is comparable with supervised \citep{christmann2016learning,wang2023tspam} and SSL models \citep{liu2018generalization}.
\end{remark}

\begin{table*}[!t]
\centering
\caption{Average Accuracy $\pm$ standard deviation (\%) on synthetic additive data for classification with fixed label percentages in each class ($r=5\%$), uninformative variable ($p_u$) and noisy variable numbers ($p_n$). (Traditional and deep SSL baselines are colored.)}
\resizebox{\textwidth}{!}{
\begin{tabular}{cccccccccccccc}
\toprule 
\multirow{2}{*}{Model} & \multicolumn{2}{c}{r = 5\%, $p_u$ = $p_n$ = 0} & \multicolumn{2}{c}{r = 5\%, $p_u$ = 10, $p_n$ = 0} & \multicolumn{2}{c}{r = 5\%, $p_u$ = 0, $p_n$ = 10} & \multicolumn{2}{c}{r = 5\%, $p_u$ = $p_n$ = 10}\\
\cmidrule(r){2-3} \cmidrule(r){4-5} \cmidrule(r){6-7} \cmidrule(r){8-9}
&  Unlabeled  &   Test
&  Unlabeled  &   Test
&  Unlabeled  &   Test
&  Unlabeled  &   Test\\
\midrule
$\ell_1$-SVM    &-     &  83.914 $\pm$ 6.410     &- &62.713 $\pm$ 6.098
& - & 62.261 $\pm$ 6.550  & - &  54.791 $\pm$ 6.951 \\
SpAM    &-     &  84.150 $\pm$ 6.104     &- &65.091 $\pm$ 5.917   & - & 64.814 $\pm$ 6.039  & - & 54.413 $\pm$ 6.295\\
CSAM  &-     &  86.597 $\pm$ 5.424     &- &69.717 $\pm$ 5.101
& - & 65.178 $\pm$ 5.255  & - &  61.980 $\pm$ 5.701 \\
TSpAM &-     &  86.993 $\pm$ 5.340     &- &71.044 $\pm$ 5.079
& - & 67.340 $\pm$ 4.959  & - &  63.145 $\pm$ 5.130 \\
\rowcolor{rowgray}
LapSVM     &88.814 $\pm$ 5.398    & 88.850 $\pm$ 5.269   &59.992 $\pm$ 5.259& 60.325 $\pm$ 5.184   & 55.630 $\pm$ 8.213&  55.957 $\pm$ 8.292  & 55.137 $\pm$ 8.414 & 55.203 $\pm$  8.496\\
\rowcolor{rowgray}
f-FME     &89.141 $\pm$ 3.172    & 89.305 $\pm$ 3.359    &64.495 $\pm$ 4.033& 64.611 $\pm$ 4.208   & 59.671 $\pm$ 6.473&  59.801 $\pm$ 6.655  & 59.311 $\pm$ 6.602 & 59.407 $\pm$  6.659\\
\rowcolor{rowgray}
AWSSL     &\textbf{91.259 $\pm$ 2.871}    & 90.211 $\pm$ 3.077    &83.691 $\pm$ 3.423& 83.950 $\pm$ 3.519   & 73.701 $\pm$ 4.105&  73.859 $\pm$ 4.322  & 72.255 $\pm$ 4.211 & 72.370 $\pm$  4.428\\
\rowcolor{rowgray}
RGL &90.422 $\pm$ 2.909     &  90.026 $\pm$ 3.477    &84.065 $\pm$ 4.501 &84.879 $\pm$ 4.711
& 77.726 $\pm$ 4.591 & 78.041 $\pm$ 4.510  & 75.155 $\pm$ 4.965 &  75.413 $\pm$ 4.708 \\
\rowcolor{rowgray}
SALE &89.717 $\pm$ 2.811     &  90.149 $\pm$ 2.665     &85.742 $\pm$ 4.132 &85.971 $\pm$ 4.018
& 79.071 $\pm$ 4.709 & 79.844 $\pm$ 4.277  & 77.201 $\pm$ 4.697 &  77.891 $\pm$ 4.431 \\
\rowcolor{rowgray2}
SSNP   &90.492 $\pm$ 3.059 
&  89.871 $\pm$ 3.218     &\textbf{86.130 $\pm$ 3.922} &85.908 $\pm$ 4.105   & 78.250 $\pm$ 4.294 & 78.062 $\pm$ 4.133  &77.462 $\pm$ 4.412 & 77.601 $\pm$ 5.513\\
\rowcolor{rowgray2}
RER   &89.416 $\pm$ 3.407    & 89.930 $\pm$ 3.622    & 85.195 $\pm$ 3.642 & 85.870 $\pm$ 3.703   & 80.933 $\pm$ 4.016 & 81.049 $\pm$ 4.055  &78.981 $\pm$ 4.302 & 79.112 $\pm$  4.517\\
S$^2$MAM-F  & 89.652 $\pm$ 3.312 & 90.015 $\pm$ 3.451 & 85.103 $\pm$ 3.526 & 85.621 $\pm$ 3.617 & 81.242 $\pm$ 3.950 & 81.351 $\pm$ 4.114 & 79.550 $\pm$ 4.210 & 79.681 $\pm$ 4.425 \\
S$^2$MAM   &89.979 $\pm$ 3.255    & \textbf{90.309 $\pm$ 3.409}    & {85.517 $\pm$ 3.481} & \textbf{86.015 $\pm$ 3.575}   & \textbf{81.702 $\pm$ 3.897} & \textbf{81.855 $\pm$ 4.055}  &\textbf{80.012 $\pm$ 4.177} & \textbf{80.112 $\pm$  4.370}\\
\bottomrule
\end{tabular}}
\label{t3-simulation-classification}
\end{table*}

\section{Experimental Evaluations}\label{section4}

This section validates the effectiveness and efficiency of the proposed S$^2$MAM across simulated and real-world high-dimensional tasks. We implement all experiments in Python. Due to space limitations, neural extensions and supplementary results on synthetic data, UCI datasets, and pixel-level noise analyses are detailed in \textit{Appendix}.

\subsection{Baselines and Experimental Setup}

\textbf{Baselines:} We compare S$^2$MAM against three branches of competitors:
\textbf{(i) Supervised Sparse Models:} $\ell_1$-SVM \citep{zhu-2003}, Lasso \citep{Tibshirani-JRSS-1994}, SpAM \citep{Ravikumar-ICNIPS-2007}, CSAM \citep{yuan2023csam}, TSpAM \citep{wang2023tspam}, and Deep Analytic Networks (DAN) \citep{dinh2020consistent}.
\textbf{(ii) Traditional SSL Approaches:} LapSVM/LapRLS \citep{belkin2006manifold}, f-FME \citep{qiu2018accelerating}, AWSSL \citep{nie2019semi}, RGL \citep{kang2019robust}, SALE \citep{nie2021adaptive}, and RER \citep{bao2024robust}.
\textbf{(iii) Deep SSL methods:} SSNP \citep{wang2022np}, SemiReward \citep{li2024semireward}, FlexMatch \citep{zhang2021flexmatch}, COREG \citep{lu2023co}, VAE \citep{cemgil2020autoencoding}, SSDKL \citep{jean2018semi}, and PLF \citep{jo2024deep}.
Notably, we evaluate the accelerated variant \textbf{S$^2$MAM-F}, which uses RFF \citep{rahimi2007random} to improve scalability. 

\begin{table*}[!t]
\centering
\caption{Average MSE $\pm$ standard deviation of 10 repeated experiments on ADNI datasets with different label percentages ($r$) and noisy variable numbers ($p_n$). Notably, noisy features are drawn from $\mathcal{N}(100,100)$. The upper and lower tables refer to prediction results for the "Fluency" and "ADAS" cognitive scores in the ADNI dataset. Notably, $^*$ highlights the interpretable baselines.}
\resizebox{0.95\textwidth}{!}{
\begin{tabular}{cccccccccccccc}
\toprule 
\multirow{2}{*}{Model} & \multicolumn{2}{c}{r = 20\%, $p_n$ = 0} & \multicolumn{2}{c}{r = 20\%, $p_n$ = 10} & \multicolumn{2}{c}{r = 50\%, $p_n$ = 0} & \multicolumn{2}{c}{r = 50\%, $p_n$= 10}\\
\cmidrule(r){2-3} \cmidrule(r){4-5} \cmidrule(r){6-7} \cmidrule(r){8-9}
&  Unlabeled  &   Test
&  Unlabeled  &   Test
&  Unlabeled  &   Test
&  Unlabeled  &   Test\\
\midrule
Lasso$^*$     &-    & 0.941 $\pm$ 0.281     &- &1.359 $\pm$ 0.733
&- &0.668 $\pm$ 0.124  & - &  0.833 $\pm$ 0.474 \\
SpAM$^*$    &-     & 0.831 $\pm$ 0.228     &- &1.266 $\pm$ 0.646   & - & 0.589 $\pm$ 0.110  &- & 0.732 $\pm$ 0.417\\
DAN    &-     &  0.794 $\pm$ 0.197     &- &1.210 $\pm$ 0.611   & - & 0.637 $\pm$ 0.105  &- & 0.793 $\pm$ 0.373\\
\rowcolor{rowgray}
LapRLS     &0.915 $\pm$ 0.301     & 0.932 $\pm$ 0.313    &1.478 $\pm$ 0.812 & 1.617 $\pm$ 0.834   &0.823 $\pm$ 0.215 &0.838 $\pm$ 0.224  &1.142 $\pm$ 0.511  &  1.167 $\pm$ 0.525\\
\rowcolor{rowgray}
RER    &0.780 $\pm$ 0.184    &0.794 $\pm$ 0.201    &0.807 $\pm$ 0.249 & 0.821 $\pm$ 0.266   & 0.437 $\pm$ 0.142  & 0.448 $\pm$ 0.157  &0.477 $\pm$ 0.225  &  0.496 $\pm$ 0.249\\
\rowcolor{rowgray2}
VAE     &0.743 $\pm$ 0.324     & 0.754 $\pm$ 0.330    &0.812 $\pm$ 0.397  & 0.825 $\pm$ 0.411   & 0.474 $\pm$ 0.115 &  0.493 $\pm$ 0.123  & 0.526 $\pm$ 0.226  &  0.541 $\pm$  0.241\\
\rowcolor{rowgray2}
COREG   &0.748 $\pm$ 0.308 & 0.761 $\pm$ 0.316  &0.984 $\pm$ 0.423   & 1.020 $\pm$ 0.434  &0.527 $\pm$ 0.276 & 0.546 $\pm$ 0.283  & 0.513 $\pm$ 0.384 &  0.531 $\pm$  0.393\\
\rowcolor{rowgray2}
SSDKL     &\textbf{0.721 $\pm$ 0.321 }   & \textbf{0.739 $\pm$ 0.337}    &0.848 $\pm$ 0.446 & 0.867 $\pm$ 0.462   &0.442 $\pm$ 0.271 &  0.454 $\pm$ 0.279 &  0.524 $\pm$  0.391 & 0.547 $\pm$  0.403\\
S$^2$MAM-F$^*$ & 0.739 $\pm$ 0.146 & 0.752 $\pm$ 0.155 & 0.795 $\pm$ 0.223 & 0.818 $\pm$ 0.235 & 0.431 $\pm$ 0.127 & 0.441 $\pm$ 0.133 & 0.477 $\pm$ 0.203 & 0.494 $\pm$ 0.211 \\
S$^2$MAM$^*$     &0.730 $\pm$ 0.133    &0.747 $\pm$ 0.147    &\textbf{0.786 $\pm$ 0.214} & \textbf{0.804 $\pm$ 0.228}   & \textbf{0.423 $\pm$ 0.119}  & \textbf{0.430 $\pm$ 0.130}  &\textbf{0.464 $\pm$ 0.196}  &  \textbf{0.483 $\pm$ 0.205}\\
\midrule
Lasso$^*$     &-    &  1.179 $\pm$ 0.376     &- &1.469 $\pm$ 0.817
&- &0.824 $\pm$ 0.255  & - &  0.961 $\pm$ 0.511 \\
SpAM$^*$    &-     &  1.250 $\pm$ 0.335     &- &1.545 $\pm$ 0.748   & - & 0.831 $\pm$ 0.217  &- & 1.017 $\pm$ 0.470\\
DAN    &-     & 1.470 $\pm$ 0.346     &- &1.844 $\pm$ 0.773   & - & 0.962 $\pm$ 0.230  &- & 1.672 $\pm$ 0.515\\
\rowcolor{rowgray}
LapRLS     &1.075 $\pm$ 0.416     & 0.973 $\pm$ 0.423    &1.813 $\pm$ 0.934 & 1.706 $\pm$ 0.945   & 0.944 $\pm$ 0.290  & 0.898 $\pm$ 0.296  &1.379 $\pm$ 0.532  &  1.409 $\pm$ 0.544\\
\rowcolor{rowgray}
RER  &0.782 $\pm$ 0.265     & 0.801 $\pm$ 0.273    &0.828 $\pm$ 0.351 & 0.817 $\pm$ 0.358   &  0.624 $\pm$ 0.228 & 0.618 $\pm$ 0.208    &0.680 $\pm$ 0.272  &  0.698 $\pm$  0.287\\
\rowcolor{rowgray2}
VAE     &0.816 $\pm$ 0.399     & 0.808 $\pm$ 0.418    &1.089 $\pm$ 0.553  & 0.924 $\pm$ 0.571   & 0.642 $\pm$ 0.253 &  0.633 $\pm$ 0.261  & 0.794 $\pm$ 0.509  &  0.760 $\pm$  0.521\\
\rowcolor{rowgray2}
COREG     &\textbf{0.766 $\pm$ 0.374}    & \textbf{0.748 $\pm$ 0.386}    &0.968 $\pm$ 0.515 & 0.735 $\pm$ 0.528   & 0.619 $\pm$ 0.277 &  0.625 $\pm$ 0.285  & 0.762 $\pm$ 0.452 &  0.736 $\pm$  0.467\\
\rowcolor{rowgray2}
SSDKL     &0.818 $\pm$ 0.383    & 0.794 $\pm$ 0.396    &0.941 $\pm$ 0.532 & 0.920 $\pm$ 0.541  &  0.617 $\pm$ 0.282 &\textbf{0.605 $\pm$ 0.269}   &  0.772 $\pm$  0.473 &  0.730 $\pm$  0.481\\
S$^2$MAM-F$^*$ & 0.778 $\pm$ 0.247 & 0.794 $\pm$ 0.262 & 0.826 $\pm$ 0.327 & 0.814 $\pm$ 0.337 & 0.625 $\pm$ 0.214 & 0.618 $\pm$ 0.197 & 0.676 $\pm$ 0.255 & 0.693 $\pm$ 0.272 \\
S$^2$MAM$^*$   &0.771 $\pm$ 0.241     & 0.783 $\pm$ 0.255    &\textbf{0.816 $\pm$ 0.321} & \textbf{0.801 $\pm$ 0.330}   &  \textbf{0.614 $\pm$ 0.204} & 0.609 $\pm$ 0.192    &\textbf{0.663 $\pm$ 0.251}  &  \textbf{0.681 $\pm$  0.266}\\
\bottomrule
\end{tabular}}
\label{t-ADNI}
\end{table*}

\begin{table*}[!t]
\caption{Extended experiments with average accuracy, standard deviation (SD), and training time cost (minutes) on COIL-20 image (upper panel) and CelebA-HQ (lower panel) data. Only $30\%$ samples of COIL and $0.5\%$ of CelebA-HQ in the training set are labeled.}
\centering
\resizebox{\textwidth}{!}{
\begin{tabular}{c c c c c 
>{\columncolor{rowgray}}c 
>{\columncolor{rowgray}}c 
>{\columncolor{rowgray}}c 
>{\columncolor{rowgray}}c 
>{\columncolor{rowgray}}c 
>{\columncolor{rowgray}}c 
>{\columncolor{rowgray2}}c 
>{\columncolor{rowgray2}}c
>{\columncolor{rowgray2}}c cc}
\toprule
Models& $\ell_1$-SVM & SpAM& CSAM& TSpAM& LapSVM & f-FME & AWSSL& RGL& SALE &RER &SSNP & FlexMatch & SemiReward & S$^2$MAM-F & S$^2$MAM\\
\midrule
Accuracy & 67.329 & 69.917& 73.577& 72.230& 81.092&  85.518 & 86.821& 83.416& 87.235 &85.219 &83.370 & 87.945 &87.518  & 87.952 & \textbf{88.211}\\
SD & 0.583 & 0.709& 0.622& 0.616& 0.417& \textbf{0.408}  & 0.430& 0.527& 0.616 &0.452 &0.429 & 0.415  &0.397  & 0.435 & 0.427\\
Time Cost & \textbf{0.2} & 0.9& 2.3& 2.5& 0.6& 1.5 & 2.7& 3.1 &4.1 & 2.2  &1.8 &9.5 &7.4 & 1.6 &2.4\\
\midrule
Accuracy & 58.412 & 60.255& 65.109& 64.882& 76.505& 81.334 & 84.110& 82.559& 88.050 &85.912 &84.440 & 91.850 & 90.625 & 91.683 & \textbf{92.104}\\
SD & 0.815 & 0.902& 0.775& 0.781& 0.655& 0.590 & 0.512& 0.605& 0.550 &0.525 &0.510 & 0.445 & 0.462 & 0.431 & \textbf{0.415}\\
Time Cost & \textbf{0.8} & 2.1& 5.5& 5.9& 1.8& 3.5 & 6.2& 7.5 & 9.8 & 5.1 & 4.5 & 35.2 & 28.6 & 5.5 & 8.4\\
\bottomrule
\end{tabular}}
\label{expimage1}
\end{table*}

\begin{table}[!t]
\centering
\caption{Regression estimation of root mean square error (RMSE) on AgeDB images, with a label ratio of $r=0.5\%$.}
\resizebox{0.48\textwidth}{!}{
\begin{tabular}{c 
>{\columncolor{rowgray2}}c
>{\columncolor{rowgray2}}c
>{\columncolor{rowgray2}}c
>{\columncolor{rowgray2}}c
cc}
\hline
Models & COREG & SSDKL & PLF & SemiReward & S$^2$MAM & S$^2$MAM-F\\
\hline
RMSE & 17.456 & 17.728 & 17.025 & \textbf{16.215} & 16.515 & 16.808 \\
SD & 2.121 & 2.305 & 1.565 & \textbf{0.650} & 0.805 & 0.870 \\
Time & 13.8 & \textbf{9.1} & 10.8 & 22.1 & 12.4 & 9.7 \\
\hline
\end{tabular}
}
\vspace{-0.5cm}
\end{table}

\textbf{Hyperparameters:} For fairness, the penalty coefficients are tuned across $[10^{-4},10^{-3},10^{-2},10^{-1}]$ via leave-one-out cross-validation, which are shared for all regularized approaches. Let $\tau_j=1$ for all $j\in[1,2,\cdots,p]$ for additive baselines \citep{wang2023tspam}. 
The bandwidth $\mu$ for similarity measure is selected within $[10^{-4},10^{-3},10^{-2},10^{-1},1]$. We repeat each experiment $100$ times and report the average accuracy and standard deviation across different data settings. 
The selection of informative feature size $C$ is stated in Appendix. The parameters for other methods were set according to the corresponding references.

\textbf{Benchmarks:} As summarized in \textit{Appendix}, 4 synthetic, 8 UCI, and 4 real-world datasets are utilized in the experiments, including the high-dimensional Alzheimer's Disease Neuroimaging Initiative (ADNI) clinical records, COIL-20 image, CelebA-HQ images \citep{lee2020maskgan}, and AgeDB images \citep{moschoglou2017agedb}. To evaluate the robustness of S$^2$MAM, $p_u$ uninformative variables in $\mathcal{N}(0,1)$ and $p_n$ noisy variables in $\mathcal{N}(100,100)$ are designed as corruptions \citep{bao2024robust}.
Due to space limitations, empirical results on additional datasets, ablation studies, and interpretable visualizations are presented in \textit{Appendix}. 

\textbf{Setup \& Metrics:} S$^2$MAM uses logistic loss for classification and squared loss for regression. Supervised baselines use only labeled data ($r$), whereas SSL methods access both labeled and unlabeled data. We report the average accuracy (classification) or Mean Squared Error (MSE/RMSE) $\pm$ standard deviation over 100 and 10 repetitions for synthetic and image datasets, respectively. 

\subsection{Experiments on Synthetic Data}

Following the experimental design in \citep{chen-2020-tnnls-spmam, wang2023tspam}, we consider the following additive discriminant function
$f^*(x_i) = (x_{i}^{(1)}-0.5)^2 + (x_{i}^{(2)}-0.5)^2 -0.08,$
where $x_{i}^{(j)}=(W_{ij}+U_i)/2$. $W_{ij}$ and $U_i$ are independently from $U(0,1)$ for $i=1, \cdots, 200$, $j=1, \cdots, 100$. 
The category label satisfies $y_i=0$ when $f(x_i) \le 0$ and $1$ otherwise.
After equally dividing the entire dataset into training and test sets, $5\%$ of the samples per class from the training set are randomly selected as the labeled set.

As presented in Table \ref{t3-simulation-classification}, both irrelevant and noisy features are harmful. Fortunately, even with irrelevant and noisy information, S$^2$MAM effectively identifies informative variables, yielding the highest prediction accuracy and the lowest variance among all baselines. Moreover, the extended visualization results in the Appendix help to demonstrate the interpretability of S$^2$MAM more effectively.

\subsection{Robustness Analysis on ADNI Clinical Records} \label{sec_adnis}

We evaluate the model's capability in high-dimensional regression using the ADNI dataset ("Fluency" and "ADAS" scores, with 326 features). 
The results in Table \ref{t-ADNI} corroborate the efficacy of S$^2$MAM, yielding an average MSE approximately 0.119 lower than that of the most competitive baselines. These results suggest that while existing shallow and deep SSL methods often struggle with uninformative dimensions that distort the graph Laplacian, the proposed S$^2$MAM successfully isolates the informative manifold structure, thereby maintaining high predictive accuracy even in sparse and noisy data environments.

\subsection{Evaluation on High-Dimensional Image Data} \label{sec_coil}

Beyond tabular data, we evaluate classification performance on the \emph{COIL-20} and \emph{CelebA-HQ} datasets, and regression performance on the \emph{AgeDB} dataset. Here, standard CNN backbones were employed to extract initial feature vectors as described in \textit{Appendix}.

The results in Table \ref{expimage1} highlight a critical trade-off between predictive accuracy and computational efficiency. While Deep SSL methods like FlexMatch and SemiReward achieve high accuracy, they incur substantial computational costs due to their complex architectures, e.g., FlexMatch requiring 35 minutes on CelebA-HQ. S$^2$MAM-F with Fourier acceleration provides a more practical, faster alternative for scenarios with limited computational resources.

Furthermore, compared to shallow and deep SSL baselines, our method shows consistent accuracy improvements. This indicates that S$^2$MAM's additive modeling capability (RMSE 16.515) outperforms the deep baseline PLF (17.025) and SSDKL (17.728), validating S$^2$MAM's efficiency and scalability across diverse learning tasks.

\section{Conclusion} \label{section5}
This paper proposes a semi-supervised meta additive model, called S$^2$MAM, to enhance the robustness and interpretability of manifold regularization in settings with redundant and noisy input variables. 
Compared with existing SSL models with manifold regularization and deep SSL models, the proposed approach achieves variable selection, interpretability, and robust estimation simultaneously. Theoretical and empirical evaluations verify its superiority over some state-of-the-art learning models. 
Future work will focus on scaling bilevel learning to handle ultra-high-dimensional scenarios.

\section*{Acknowledgement}

We sincerely thank Feng Zheng and Zeyu Zhang for providing some valuable experimental suggestions on our prior construction of this manuscript.

\section*{Contribution Statement}
Xuelin Zhang was responsible for writing the original manuscript, conducting the experiments, and handling the revisions. Hong Chen acted as the project administrator, conceptualized the core ideas, and contributed to manuscript revisions. Yingjie Wang contributed to the algorithm design of the additive models and provided the source code for the baseline methods. Tieliang Gong provided guidance on the statistical theory regarding convergence and generalization. Bin Gu provided guidance on the design of the bi-level optimization algorithm. Explicitly,  Feng Zheng and Zeyu Zhang also helped to offer suggestions for the experimental setup on our prior construction of this manuscript.

\section*{Impact Statement}
This paper presents work aimed at advancing the field of Semi-Supervised Learning and Interpretable Machine Learning. We believe this work can deepen our understanding of robust graph Laplacian construction in the presence of noisy, high-dimensional data and widen the applications of bilevel optimization for simultaneous variable selection and transparent model estimation. 

The proposed S$^2$MAM framework significantly enhances interpretability and resilience against feature corruption compared to classical and recent deep semi-supervised methods. There may be some potential societal consequences of our work, none of which we feel must be specifically highlighted here.

\bibliography{icml2026}
\bibliographystyle{icml2026}

\end{document}